%% file: MAIN.tex
\documentclass[manuscript,table]{acmart}
\usepackage{booktabs}
\usepackage{caption}
\usepackage{amsmath}
\usepackage{makecell}
\usepackage{array,subcaption}

\usepackage{tabto}

\input{COMMANDS}

\begin{document}

\title{LLMs as Research Tools: A Large Scale Survey of Researchers’ Usage and Perceptions}

\author{Zhehui Liao}
\affiliation{%
  \institution{University of Washington}
  \city{Seattle}
  \country{US}
}
\author{Maria Antoniak}
\affiliation{%
  \institution{Pioneer Centre for Artificial Intelligence, University of Copenhagen}
  \city{Copenhagen}
  \country{Denmark}
}
\author{Inyoung Cheong}
\affiliation{%
  \institution{Princeton University}
  \city{New York}
  \country{US}
}
\author{Evie Yu-Yen Cheng}
\affiliation{%
  \institution{Allen Institute for Artificial Intelligence (Ai2)}
  \city{Seattle}
  \country{US}
}
\author{Ai-Heng Lee}
\affiliation{%
  \institution{Allen Institute for Artificial Intelligence (Ai2)}
  \city{Seattle}
  \country{US}
}
\author{Kyle Lo}
\affiliation{%
  \institution{Allen Institute for Artificial Intelligence (Ai2)}
  \city{Seattle}
  \country{US}
}
\author{Joseph Chee Chang}
\affiliation{%
  \institution{Allen Institute for Artificial Intelligence (Ai2)}
  \city{Seattle}
  \country{US}
}
\author{Amy X. Zhang}
\affiliation{%
  \institution{University of Washington}
  \city{Seattle}
  \country{US}
}

\renewcommand{\shortauthors}{}

\begin{abstract}
The rise of large language models (LLMs) has led many researchers to consider their usage for scientific work.
Some have found benefits using LLMs to augment or automate aspects of their research pipeline, while others have urged caution due to risks and ethical concerns.
Yet little work has sought to quantify and characterize how researchers use LLMs and why.
We present the first large-scale survey of 816 verified research article authors to understand how the research community leverages and perceives LLMs as research tools.
We examine participants' self-reported LLM usage, finding that 81\% of researchers have already incorporated LLMs into different aspects of their research workflow.
We also find that traditionally disadvantaged groups in academia (non-White, junior, and non-native English speaking researchers) report higher LLM usage and perceived benefits, suggesting potential for improved research equity.
However, women, non-binary, and senior researchers have greater ethical concerns, potentially hindering adoption.
\end{abstract}


\maketitle

\section{Introduction}

From Vannevar Bush's hypothesized \emph{Memex} in 1945~\cite{bush1945we} to Apple's vision of the \emph{Knowledge Navigator} in 1987~\cite{newendorp2024apple}, many have long envisioned tools that  \emph{organize and interact with the sum of our knowledge}  to enable us to better conduct complex knowledge work and push the boundaries of what we know.
Recent advancements in generative AI technologies that are trained on terabytes of text scraped from the internet~\cite{soldaini2024dolma} and fine-tuned to provide a chat interface~\cite{ouyang2022traininglanguagemodelsfollow} bring us one step closer to making this dream a reality.
More directly, this technology has also sparked a surge of interest across research, industry funding, and startups
to develop a new generation of research support tools \cite{lee2024paperweaver,elicit,taylor2022galactica,gero2022sparks,sun2024metawriter,undermind,consensus}.

In fact, the recent burst in popularity of widely available generative AI tools, such as ChatGPT,\footnote{The ChatGPT mobile app has more than 100 million downloads on the Android app store and more than 1 million ratings on the iPhone app store, as of September 2024.} and findings from small-scale interview and survey studies with researchers~\cite{morris2023scientists,Fecher2023FriendOF} suggest that many in the research community have already found benefits in incorporating current generative AI models into their research workflows.
Adopting this new generation of knowledge tools has opened up many possibilities, such as improved efficiency, greater research equity, and inspiring novel ideas.
At the same time, new tools both breathe new life into familiar research risks and ethical concerns --- like transparency, reproducibility, plagiarism, and data fabrication --- while introducing new dangers to the research process.
It is possible that these tools could result in researchers losing essential skills \citep{Kobiella2024if, ahmad2023impact}, the development of emerging social norms and associated reputational costs \citep{hosseini2023ethics}, and a decrease in research creativity, among other possibilities.
Differences in perceptions about risks, ethics, and social acceptability across demographic groups and researcher backgrounds could also drive differences in adoption, so that any benefits accrue unevenly. This could potentially exacerbate existing structural barriers in academia due to biases and other factors~\cite{Goyes2023RichSP}.

While prior work has mostly focused on research domain-specific investigations~\cite{kapania2024m,Liang2024MappingTI,Russo2024TheAR,Ke2024ExploringTF,Williams2023AlgorithmicGI} or small-scaled surveys or qualitative interviews~\cite{morris2023scientists,Fecher2023FriendOF}, we conducted a large scale survey with verified published authors.
These authors were sourced from \platform{}, a platform that maintains an open repository of published researchers from a wide range of research domains, demographic backgrounds, and research experiences.
Our survey of these researchers was designed to explore the following research questions.
\begin{itemize}
\item \textbf{RQ1}: What are the different ways researchers \textbf{use}\footnote{Usages of LLMs in the research process that we examine: information seeking, editing, ideation and framing, direct writing, data cleaning and analysis, data generation} LLMs in their research process today?
\item \textbf{RQ2}: How does the \textbf{background}\footnote{Researcher background characteristics that we examine: race, gender, native English speaker, research experience, field of research} of a researcher relate to the way they \textbf{use} LLMs?
\item \textbf{RQ3}: What are researchers' \textbf{perceptions}\footnote{Researcher perceptions of LLMs that we examine: risks, benefits, ethics, and willingness to disclose to peers and reviewers} of LLMs for \textbf{usage} in different parts of the research process?
\item \textbf{RQ4}: How does the way a researcher \textbf{uses} LLMs relate to their \textbf{perceptions}?
\item \textbf{RQ5}: How does the \textbf{background} of a researcher relate to their \textbf{perceptions}?
\item \textbf{RQ6}: How does the \textbf{source}\footnote{Sources of LLMs that we examine: non-profit vs. for-profit entities} of an LLM affect researchers' \textbf{perceptions} and \textbf{usage}?
\end{itemize}

Our survey focused on better understanding how researchers are \textit{actually} using LLM-based researcher tools in their own work \textit{today}, and how they perceive the risks and benefits of leveraging LLMs for different research tasks.
In particular, we were also interested in researchers' perceptions not only of LLM usage but also about the acceptability of using these tools, and the possible differences in perception across demographic groups, leading us to recruit researchers across nationalities, languages, career stages, discipline, gender, age, etc.
The differences we uncover between these groups reveal rapidly changing social norms around the usage of AI tools in research, highlighting important considerations around research equity and broader adoption.

In particular, we find around 81\% of researchers we surveyed have used LLMs in one or more places in their research pipeline, with the tasks of Information Seeking and Editing reported most frequently and Data Analysis and Generation reported least frequently.
We also find surprising differences between researchers of different demographic groups.
Based on self-reporting, we found that researchers who are non-White, non-native English speaking, and junior researchers both use LLMs more frequently and also perceive higher benefits and lower risks. As people with these demographics traditionally tend to face certain structural barriers, our findings suggest that LLMs can help with improving research equity.
Meanwhile, women and non-binary researchers have greater ethical concerns, as do those with more years of research experience.
In addition, while LLMs are broadly used across all fields, we see significantly greater comfort with disclosure of usage in computer science fields as well as lower ethical concerns compared to other disciplines.
This suggests that to truly achieve broad  adoption, ethical issues around LLMs must be confronted and social norms established within each field.
Finally, we find that researchers overall prefer to use LLMs from open source/non-profit entities over for-profit entities due to a variety of concerns with models from the existing major for-profit corporations.

Our contributions include the following.
\begin{itemize}
\item Large-scale survey results from 816 verified research paper authors, which we release to the public.\footnote{Link to survey results: \url{https://github.com/allenai/llm-research-survey}. All personally identifiable information in the responses, including all free-text responses have been removed. The study is approved by our university IRB, and the participants each gave explicit consent at the start of the survey for their data to be released.}
\item Detailed quantitative and qualitative analyses of the survey results, revealing how researchers currently incorporate LLMs into their workflow, and how they perceive their risks and benefits.
\item Insights into LLM usage patterns and perceptions across demographic groups, revealing
that while some traditionally disadvantaged groups in academia (non-White, non-native English speaking, and junior researchers) report higher usage and perceived benefits, other groups (women, non-binary, and senior researchers with more experience) express greater ethical concerns.
\item Lessons about researchers' preferences for non-profit and open-source models versus commercial models, along with their rationales, which can inform future development and adoption strategies for LLMs in academia.
\end{itemize}

\section{Related Work}

\subsection{LLMs as Research Support Tools: Current Practices and Benefits}

Recent work has suggested that researchers have already begun to adopt LLMs into their research workflows across various disciplines. In early 2023, Morris~\cite{morris2023scientists} conducted in-depth interviews with 20 researchers of diverse backgrounds,\footnote{Notably, this study intentionally only covered researchers from non-computer science fields.} focusing on the opportunities and potential concerns around the use of LLMs as research tools in their respective fields. This study, with many others, reveals that researchers across numerous disciplines are incorporating LLMs into different stages of research, including ideation, literature review, data creation, cleaning and analysis, programming, and, most commonly, writing or drafting research papers~\cite{Gruda_2024, than2024updating, veselovsky2023generating, ziems-etal-2024-large, bail2024can, Manning2024AutomatedSS}.

Moreover, LLMs have been explored as a potential solution to challenges in the academic publishing process.
Alvarez~\cite{alvarez2024science} suggests that LLMs could address the ``overtaxed peer-review system'', a sentiment supported by Liang et al.~\cite{liang2024feedback}, who found that over 80\% of researchers considered ChatGPT-generated feedback more beneficial than feedback from at least some human reviewers. Separately, Koller et al.~\cite{koller2023we} advocated for the use of LLMs in conference submissions, arguing that these tools help researchers ``contextualize their work, democratize knowledge, enhance data analysis, and produce better scientific output.''  As a result, the increasing adoption of LLM assistance is evident in both scientific research articles ~\cite{Liang2024MappingTI} and peer reviews~\cite{Russo2024TheAR}.

Beyond evidence of the research community adopting widely available LLM-based tools into their workflow, the AI and HCI research community have also devoted many resources in recent years to exploring the next generation of research tools powered by LLMs. Research has focused on all stages of the research workflow, from research ideation \cite{gero2022sparks,10.1145/3613904.3642698,wang-etal-2024-scimon}, paper reading \cite{lo2023semantic,fok2023qlarify}, literature review \cite{lee2024paperweaver,kang2023synergi,hsu2024chimellmassistedhierarchicalorganization}, writing \cite{kim2023metaphorian,long2023tweetorial, Gruda_2024}, peer review support \cite{sun2024metawriter,d2024marg}, and more.
Anecdotally, there has also been increased commercial interest in building LLM-based research support tools. For example, Meta released Galactica, an LLM model trained for general scientific tasks in 2022~\cite{taylor2022galactica}; Undermind allows users to converse with a chatbot that has access to research paper search results~\cite{undermind}; Elicit allows users build research paper comparison tables by extracting information across many papers~\cite{elicit}; and Consensus answers scientific questions by gathers confirming and opposing evidence across papers~\cite{consensus}. More recently, there have also been attempts to build fully autonomous end-to-end research agents based on LLMs. For example, Sakana AI built ``The AI Scientist'', which aimed to conduct research in machine learning, from generating research ideas to executing experiments, to writing the paper, and reviewing the paper~\cite{ai-scientists, castelvecchiresearchers}; and, similarly, FutureHouse launched a 10-year mission to automate biology research at scale \cite{futurehouse}.

Given this trend, there have been several recent work focused on a better understanding of how researchers leverage LLMs in their research workflow and the risks and benefits of doing so. Yet, many only focused on one specific research domain (e.g., HCI~\cite{kapania2024m}, psychology~\cite{Ke2024ExploringTF}, machine learning~\cite{Russo2024TheAR}, and management research~\cite{Williams2023AlgorithmicGI}). More closely related to our work, there has also been a small scale survey (N=72) \cite{Fecher2023FriendOF} and interview study (N=20) \cite{morris2023scientists} that covered researchers across different domains. %
Our work represents the first large-scale survey (N=816) focusing on researchers' use of LLMs across various disciplines. It provides empirical evidence for trends previously only speculated about in smaller qualitative studies~\cite[e.g.,][]{morris2023scientists} or inferred from textual analysis of published papers~\cite[e.g.,][]{Kobak_2024}.

\subsection{Risks and Ethical Implications of LLMs in Research}

While LLMs have shown great promise for a future where novel AI capabilities can have significant positive impacts on science, the current implementations have faced many critiques from the community, including claims that LLMs have a popularity bias, contain a reductive view on how researchers learn their knowledge, and generate made-up articles completely~\cite{heaven2022why, bail2024can}.

\subsubsection{Lack of precision or ``hallucination''}
One of the primary concerns of the applications of LLMs to scientific research is their insufficient level of precision and accuracy~\cite{alvarez2024science}. LLMs have been observed to generate plausible-sounding but entirely fictional content, a phenomenon popularly referred to as ``hallucination''~\cite{alkaissi2023artificial}.
For example, Galactica, an LLM trained on scientific papers~\cite{taylor2022galactica}, was taken down after producing convincing but false scientific articles~\cite{heaven2022why}. Current LLMs struggle with tasks requiring precise calculations and logical reasoning~\cite{Srivastava_Malik_Gupta_Ganu_Roth_2024}. In software development, studies indicate that developers often reject LLMs' initial code suggestions~\cite{ziegler2022productivity} and face difficulties in understanding and debugging the generated code~\cite{liang2024large, vaithilingam2022expectation, liu204noneed}. Without proper vetting, inaccurate content could contribute to the spread of misinformation and erode trust in research~\cite{Kobiella2024if, Gruda_2024}. In fields like medicine or engineering, where precision is crucial, LLM inaccuracies could have serious real-world consequences~\cite{Antoniak_Naik_Alvarado_Wang_Chen_2024, Toma2023generative}. Our results showed a similar phenomenon in using LLMs for science, where hallucination and misinformation
were among the most frequently mentioned risks by our participants based on qualitative responses.

\subsubsection{Undermined research integrity}
The adoption of LLMs in academic research raises fundamental questions about diminished research integrity and originality. Researchers have expressed worry about the potential ``proliferation of low-quality research''~\cite{bail2024can}, as LLMs may facilitate the mass production of superficial or derivative work. This concern is corroborated by Kobiella et al.~\cite{Kobiella2024if}, who found that knowledge workers experienced a decreased sense of achievement when using LLMs, driven by reduced ownership, lack of challenge, and concerns about output quality. The National Institutes of Health (NIH) has taken a strong stance against using LLMs for grant applications or reviews~\cite{nih}, cautioning that such practices undermine ``the originality of thought'' and lead to homogenization of ideas or even research misconduct~\cite{Lauer_Constant_Wernimont_2023}. Furthermore, the integration of LLMs into review writing may exacerbate existing issues of unpredictability and unfairness. Latona et al.~\cite{Russo2024TheAR} identified ``AI Review Lottery'' at ICLR 2024 where papers receiving AI-assisted reviews were more likely to be accepted, raising questions about the reliability of the peer review system.

\subsubsection{Unexplainability and obscurity}
The complexity of LLMs makes it difficult to understand or explain why they come to certain output, which in turn affects the reliability and interpretation of research results assisted by LLMs~\cite{Singh_Inala_Galley_Caruana_Gao_2024}. The obscurity issue restricts access to model's internal workings and training data, thereby hindering research transparency and verifiability~\cite{Wulff2024open}. This issue is particularly concerning because many researchers, particularly those lacking technical expertise or computational resources predominantly rely on commercial closed models~\cite{Wulff2024open, Toma2023generative}. Sallou et al. \cite{Sallou_Durieux_Panichella_2024} asserts that software engineering research faces threats to validity from the prevalent use of closed-source models, potential data leakage and reproducibility issues due to output variability and time-based drift, all of which can compromise the reliability and generalizability of research findings. While open models offer greater transparency, compared to closed models, by offering access to code and weights, many still fall short of full disclosure, commonly withholding elements like training datasets or fine-tuning processes~\cite{lisenfeld2024rethinking, Kapoo2024open}. In this work, we ask survey participants to discuss their preferences for the predominantly closed models offered by industry versus open source and non-profit alternatives.

\subsection{Demographic Influences on LLM Perception and Adoption}

In addition to the high-level benefits and risks of LLMs, individual perceptions and usage patterns of LLMs are shaped by various factors, including personality traits, age, gender, and educational background~\citep{jakesch2022different}.
For instance, in the realm of personality and age, research has shown that people with a high level of agreeability and younger people tend to have more positive views of AI, while those susceptible to conspiracy theories often have more negative perceptions~\cite{Stein2024AttitudesTA}. A notable gender gap has been observed in LLM adoption, with male users outnumbering female users, which could be mitigated through technology-related education~\cite{draxler2023genderagetechnologyeducation}. In the field of scientific research, structural biases have long led to disparities in academic publishing, citations, and career advancement along the lines of gender, race, economic status, and more~\cite{Hopkins2013DisparitiesIP, Goyes2023RichSP}. Despite some progress in recent decades, projections indicate that gender gaps in STEMM (Science, Technology, Engineering, Mathematics, and Medicine) fields may persist for generations without significant systemic reform~\cite{Holman}. Interestingly, LLMs present an opportunity to reduce certain inequities in research and publishing. They can lower barriers for non-native English speakers~\cite{morris2023scientists} and provide high-quality reviews to novice researchers who may struggle to obtain feedback from peers~\cite{chamoun2024automated}.

While LLMs show promises and challenges in academic settings, a significant gap exists in our understanding of their impact across diverse demographic groups. Quantitative research examining how LLM usage patterns and perceptions vary among different populations within academia is notably scarce. To address this gap, we conducted a large-scale survey of researchers from a wide range of backgrounds. Our aim was to explore the nuanced and potentially differential impacts of LLMs on various demographic groups within the academic community. This approach recognizes that understanding the risks and opportunities presented by LLMs is important not only for the research community as a whole but also for specific demographic subgroups who may experience unique challenges or benefits. Our findings indicate that LLMs have a disproportionate effect on researchers with different identities, suggesting both challenges and opportunities for using these tools to address longstanding inequities in the research ecosystem.

\section{Methods}

Drawing insights from prior literature, we designed a questionnaire to study researchers' usage and perception of LLMs, and recruited participants among verified published authors. We initially collected 1,226 responses and ended up with 816 responses after filtering to ensure completeness and quality. Different from prior work that only reported participants' fields of study in small-scale surveys and interviews~\cite{morris2023scientists,Fecher2023FriendOF}, we additionally collected fine-grained demographic information in our survey.
We transformed the dataset to generate the final demographic groups in which some of the response options were grouped to form coarser buckets (e.g., years of research experience), and free responses (e.g., fields of study)  were manually coded and discretized for analysis.
The survey collected both multiple-choice responses and free-text responses.
We used linear mixed-effects models to test the relationships between researchers' LLM usage and perception, as well as between researchers' background and usage and perception. For free-text responses, we conducted an iterative open thematic analysis to gain a deeper qualitative understanding of participants' perceptions of LLMs~\citep{Boyatzis1998TransformingQI,Connelly2013GroundedT}.

\subsection{Survey Design, Participant Recruitment, and Data Collection}

\subsubsection{Design and recruitment.}
When designing the questionnaire in the survey, we used the four following approaches: First, for inspiration, we looked to recent literature on using LLMs as a productivity tool for research~\cite{morris2023scientists,messeri2024artificial,bail2024can,Russo2024TheAR,wang2023human} and other scenarios~\cite{liang2024large,liu2022will}, which included qualitative interviews and survey results.
Second, we reviewed historic papers on how the research community had adopted new tools in the past, specifically around the use of crowdsourcing for data collection, user studies, and other productivity tasks~\cite{law2017crowdsourcing,kittur2008crowdsourcing,kittur2013future}.
Third, we publicized an anonymous formative survey on X/Twitter targeted towards researchers with open-ended questions about whether and how they use LLMs for research in order to help define initial categories of usage that we later refined.
Finally, we shared early drafts of the questionnaire with other researchers in our own institutions for feedback and iteration.
In the end, we classified LLM usage for research into six broad categories, each with more specific use cases under them: information seeking, editing, ideation \& framing, directing writing, data cleaning \& analysis, and data generation.
We provide the full set of final survey questions in the Supplementary Materials.

\subsubsection{Data collection}
We collected survey responses from participants who have published at least one research paper in the past.
To ensure participants were published authors, we partnered with \platform{} for targeted recruitment of researchers who are listed as an
author of at least one published paper on the platform.
\platform{} maintains a large-scale academic knowledge graph of researchers (i.e., \emph{authors}) and papers, and provides a freely available web service to browse them as \emph{author profile pages}.\footnote{\url{https://www.semanticscholar.org/}}
Researchers can \emph{claim} their author profile pages and send corrections to \platform{}, which in turn employs a quality assurance team for verifying the claims and corrections. A survey recruitment email was sent to 107,346 verified claimed authors, and the click-through rate was around 1.6\%. After click-through, 71.6\% of the participants signed the consent form to start the survey, of which 60.6\% completed the survey.
We collected 1,226 unfiltered survey responses, which we subsequently filtered to exclude those from participants who did not progress past the first page or spent fewer than 2 seconds on each question.
\textbf{In total, this resulted in $n=$ 816 survey responses that we used for our analysis}. The survey contained a mix of optional and required questions. For example, participants could choose not to disclose their demographic information, such as gender or race.
The study was reviewed and exempted by the University of Washington IRB.

\subsubsection{Limitations}
\label{limitations}

Since we recruited from verified authors listed on \platform{}, we could have tied the survey responses to participants' author metadata from \platform{} to obtain high-precision demographic information (such as a list of publications, years of experiences, institutions, pronouns, etc.).
However, for privacy concerns, we only used their email addresses for targeted recruitment of verified published authors.
We instead relied on self-reporting using optional survey questions for demographic information, and did not tie survey responses to their author metadata. While \platform{} covers a wide range of fields of study, we did find more participants to be in the field of computer science (40\%), but other fields such as social sciences, biology, medicine, and natural sciences were also represented.
There were also more men who responded to the survey (79\%), which may partly reflect the existing imbalances in these fields of study.
The detailed distributions are reported in \S\ref{section:results}.
Finally, the survey responses were collected in batches of recruitment emails over a six-month period from November 2023 to April 2024, with the bulk of the responses received in January 2024.
The uses and perceptions of researchers may change over time as LLM tools continue to evolve, but we hope this survey can give the readers a snapshot of the current state of the community and support informed decisions as we continue to build consensus and norms around the use of LLMs for research.

\subsection{Quantitative Analysis of Survey Responses}

\subsubsection{Preparing the demographic groups}
Out of the 816 responses, 644 provided demographic information. We focused on five demographic categories collected in the survey for later analysis: \textbf{gender, race, years of research experience, native language, and field of study}. The first four were collected as answers to multiple-choice questions and further consolidated into broader categories during analysis.
For example, to balance our analysis given a large proportion of men participants, we collapsed all responses from women, non-binary, and other participants into a single category.
Field of study was collected as free response and manually classified into four categories by the authors (more details on this process can be found in Appendix \ref{appendix:creating-field}).
Answers that did not fit into any categories, such as ``Prefer Not to Answer'' or ``Prefer to Self-Describe'' were filtered out from demographic-specific analysis.
We ended up with 611 responses with gender identity, 527 with racial identity, 644 with years of research experience and native language information, and 635 with field of study information.
The final distributions of demographic groups are:
\begin{itemize}
    \item \textbf{Gender}: Man (79\%); Woman, Non-Binary, Other (21\%)
    \item \textbf{Race}: White (61\%); Non-White (39\%)
    \item \textbf{Years of Research Experience}: 11+ (57\%); 4-10 (32\%); 0-3 (11\%)
    \item \textbf{Native Language}: Native English (62\%); Non-Native English (38\%)
    \item \textbf{Field of Study}: Computer Science (40\%); Social Science \& Humanities (24\%); Natural Science \& Engineering (21\%); Biology \& Medicine (15\%)
\end{itemize}

\input{tables/chi_results_subfigures}

Finally, we inspected our demographic data for correlation between certain demographic groups. For example, are a majority of the male participants also white?
Given the demographic variables are categorical, we conduct a series of Chi-square tests of independence in \texttt{R} (\texttt{chisq.test}) between all pairs of the five demographic groups, with multiple comparisons $p$-value correction using Holm-Bonferroni (\texttt{p.adjust}). Results of the Chi-square tests can be found in Table~\ref{tab:chisq-demographic-p-values} in Appendix~\ref{appendix:survey-stats}.\footnote{We previously also had another variable representing researcher experience---number of publications---but found that it was highly correlated with Years of research experience (Chisq Independence test; p-value = 3.3E-15).}

We found that most variables appear independent, except for three pairs with significant $p$-values: race and years of experience, gender and field of study, and years of experience and field of study.
In Table~\ref{tab:chi_results_subfigs}, we present contingency tables for these three demographic pairs alongside a fourth pair, race and gender, which appear un-associated.
Interpreting these results, our dataset appears to have: (1) a high proportion of more senior white researchers; (2) more researchers who identified as men in the Natural Science \& Engineering field, and (3) fewer senior researchers in the Computer Science field.
These could be a combination of sampling bias or existing imbalance in these respective fields, as discussed in the limitations (section \ref{limitations}).

\subsubsection{Statistical methods}

\begin{figure}[h!]
    \centering
    \includegraphics[width=0.75\textwidth]{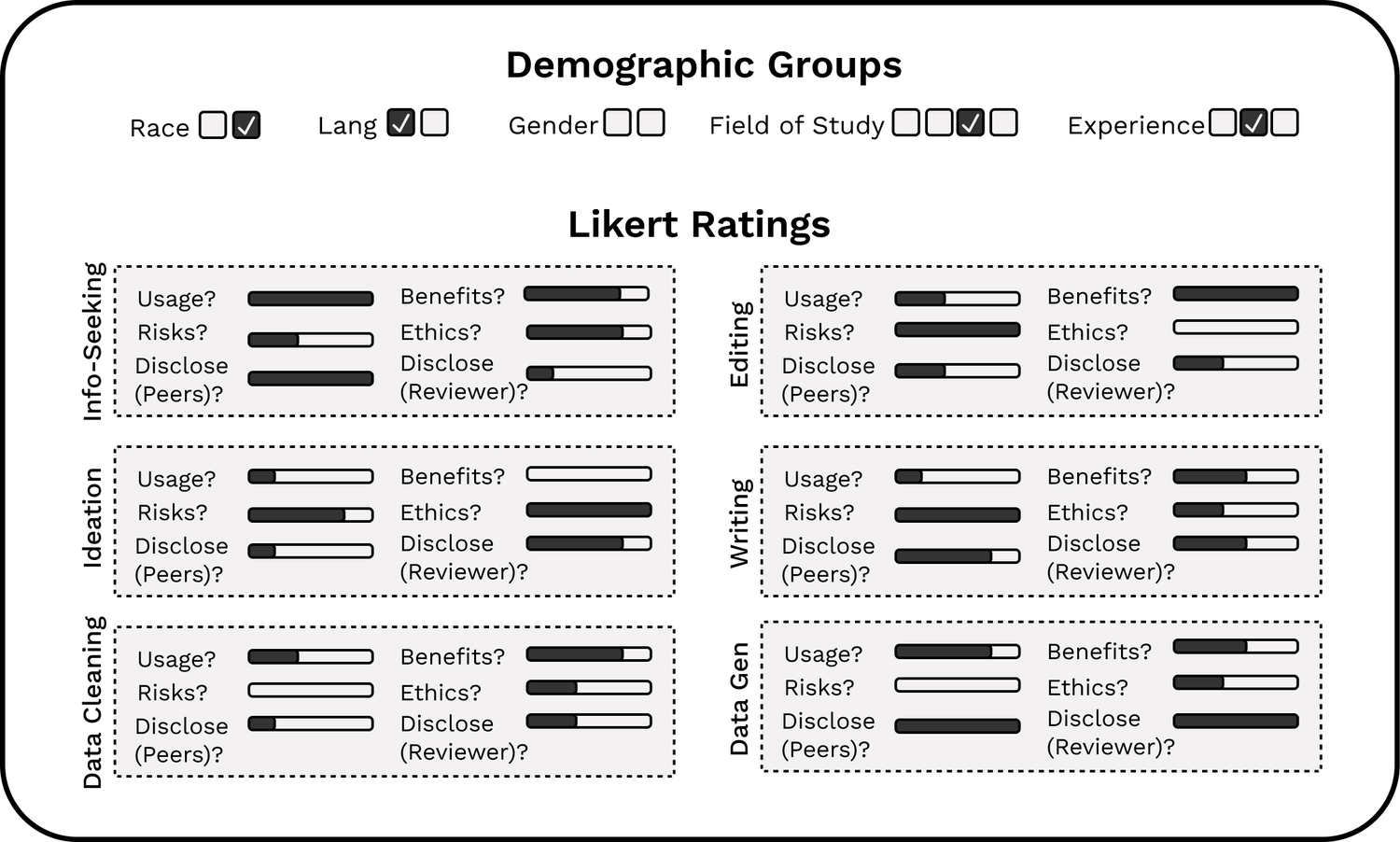}
    \caption{An example of data we collected per participant from sections of the survey that relate to our statistical modeling. Each participant provides answers to (up to) five demographic questions and (up to) 36 Likert ratings in response to questions about LLM usage frequency, perceptions, and usage types. Participants are not required to report on every question. In this example, the gender information is missing from the participant.}
    \label{figure:data-diagram}
\end{figure}

Figure~\ref{figure:data-diagram} provides a simplified diagram of an example of how our data looks per participant after completing all filtering and transformations. Each participant is labeled with (up to) five demographic categories. Each participant contributes (up to) 36 Likert ratings (an LLM Usage Frequency question and five LLM Perception questions, each repeated for six LLM Usage Types).

To address potentially correlated measurements arising from the same participants contributing multiple ratings, known as \emph{repeated measures}, we employ linear mixed effects models to test the association between participant ratings (e.g., LLM usage frequency or perceptions) and participant demographic fixed-effects, while controlling for participant-specific effects. Such models are widely used in medicine~\citep{Cnaan1997UsingTG} and behavioral sciences~\citep{Cudeck1996MixedeffectsMI} for regression analysis with repeated measures and have also seen adoption in HCI research~\citep{Hearst2019AnEO, scholarphi, paperplain, chang2023citesee}. In this work, we primarily use linear models of the form:

\begin{equation}
\text{Rating} \sim \text{Demographic} + \text{UsageType} + (1|\text{ParticipantID})
\end{equation}

For example, to measure the association between race and LLM usage, we regress participant usage (Rating) onto the race binary variable (Demographic), including additional control terms for the type of LLM usage (UsageType) as well as a random intercept term (1|ParticipantID) to capture participant-specific effects, such as when an individual has a tendency to give systematically higher or lower ratings. We can repeat this process for other demographic variables, for example swapping out race for gender or years of experience, to obtain different linear model fits.\footnote{In practice, we always attempt a second model fit that includes an \emph{interaction term} between Usage Type and Demographic. We conduct a likelihood ratio test using \texttt{anova()} in \texttt{R} between each model against a \emph{null} model which has no demographic variable (i.e., the null hypothesis of no demographic effects) and choose the interaction model if the interaction term is statistically significant; otherwise, we defer to the simpler model without an interaction term.  In our analysis, we find that rarely is the interaction term needed.}
We fit linear mixed effects models using \texttt{lme4} in \texttt{R}, which also gives us $p$-values for significance tests on the estimated fixed effects associated with each demographic variable; we correct  $p$-values for multiple comparisons across all these model fits using Holm-Bonferroni with \texttt{p.adjust}.
Each of these models was fit on thousands of participant ratings, even accounting for missing data removal due to survey answers like ``Unsure/Don't Know'' and ``I haven't used an LLM for this type of activity''; the final number of answers used in these regressions can be found in Table \ref{tab:data-overview} in Appendix \ref{appendix:survey-stats}.
The significance tests from this analysis are used to address RQ2 (\S\ref{sec:rq2}; Rating is for LLM Usage) and RQ3 (\S\ref{sec:rq3}; Rating is for LLM Perception).
For RQ4 (\S\ref{sec:rq4}), we still fit a linear mixed effects model but use LLM Usage as the response variable and replace Demographic with LLM Perception.

While linear mixed effects models can help us test whether a given demographic is significantly associated with higher/lower LLM Usage or a certain LLM perception, we  also want to know how individual \emph{levels} within each demographic variable (e.g., White versus non-White) relate to the response variable.
To do this, we conduct post hoc analyses of any significant model fit using \texttt{emmeans()} in \texttt{R} to measure and test pairwise differences in average Rating between levels in a Demographic variable holding all other variables (e.g., UsageType) constant.
These pairwise comparison tests are used to address RQ5 (\S\ref{sec:rq5}).

\subsection{Qualitative Analysis of Free-Text Responses}

To collect deeper insights beyond pre-defined multiple choices, several of the questions in our survey (Q64, 65, and 66) were paired with an optional free-text response question in which participants could elaborate and provide the reasoning behind their multiple-choice answers.
To analyze these responses, we followed an iterative open thematic analysis approach \citep{Boyatzis1998TransformingQI,Connelly2013GroundedT}.
Specifically, three of the paper authors read through and coded the same subset of the responses into thematic categories independently.
Then, the three authors meet with the entire research team to compare their codes and discuss their findings to settle on a final set of themes.
Using the final set of themes, the three authors coded the full of responses independently without replications (each response was annotated by one author).
The full set of themes (one set per question), with definitions and examples, can be found in Appendix \ref{appendix:qual-themes}.

\begin{figure}
    \centering
    \includegraphics[width=\textwidth]{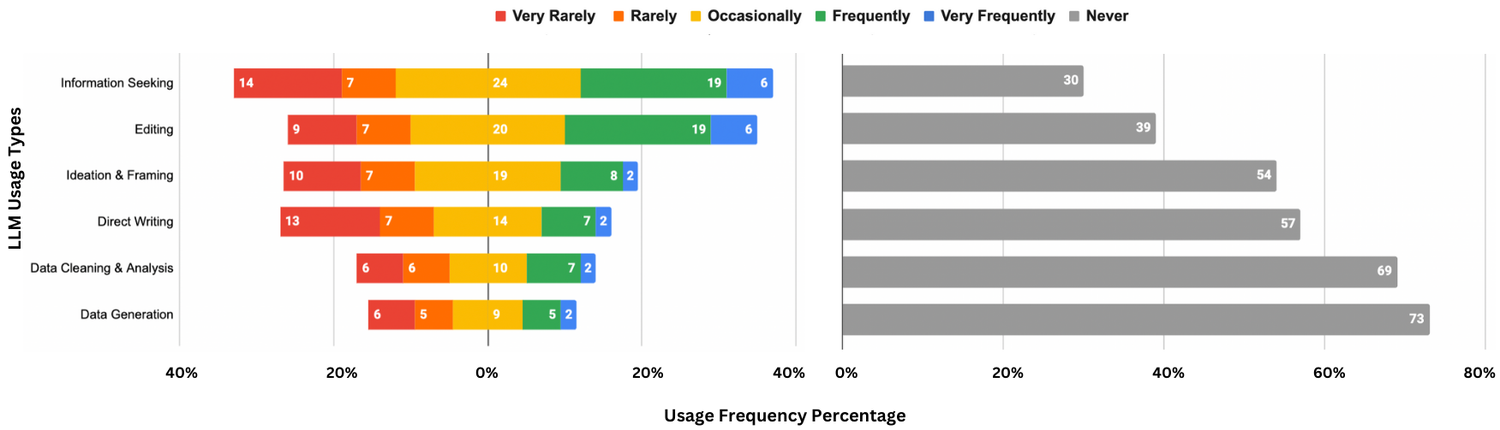}
    \caption{\textbf{Overview of Usage Frequency Divided by LLM Usage Type (N=816).} The left diverging bar chart displays the distribution of usage frequency across different types of LLM usage, with each type represented by a separate row. The frequency levels, from left to right, are: Very Rarely, Rarely, Occasionally, Frequently, and Very Frequently, with the midpoint of the chart centered at "Occasionally." The grey bar chart on the right indicates the percentage of responses that report "Never" using LLMs for each corresponding type.
    From this plot, we can tell that researchers report using LLMs for Information Seeking and Editing most frequently, and for Data Cleaning \& Analysis and Data Generation the least frequently. }
    \label{figure:usage-frequency}
\end{figure}

\begin{figure}
    \centering
    \includegraphics[width=\textwidth]{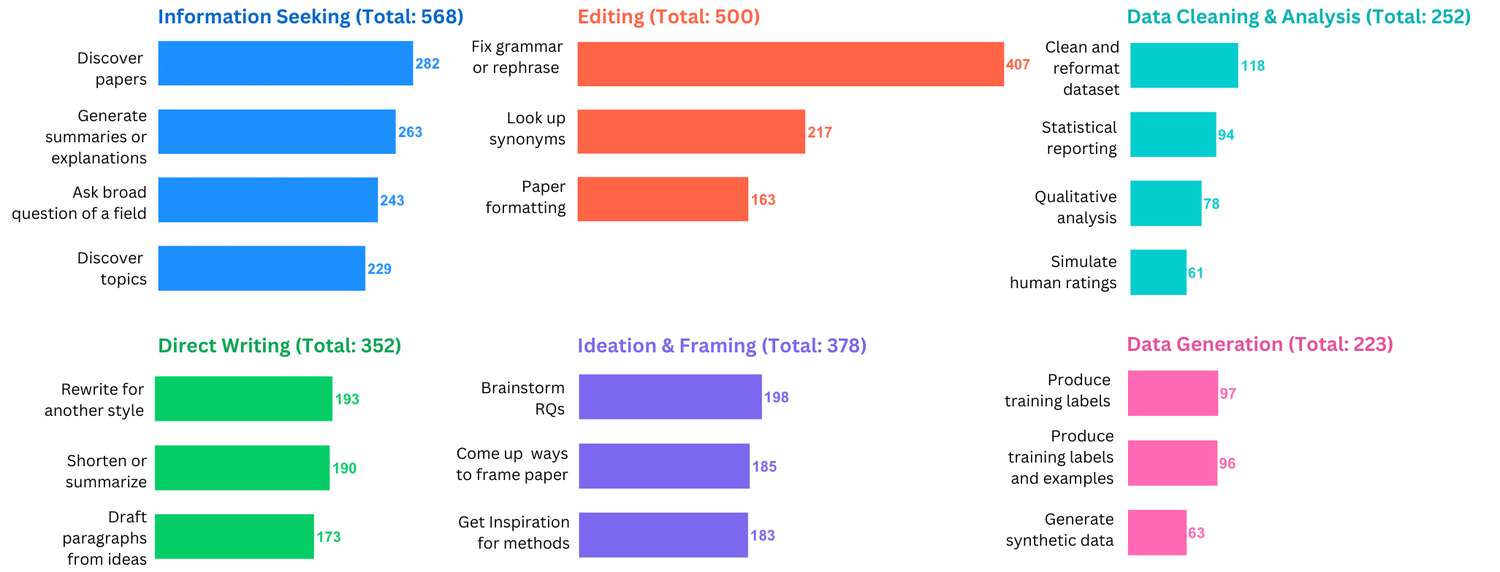}
    \caption{\textbf{LLM Usage Breakdown Under Each Usage Type (N=816).} Each bar chart shows the number of participants who reported using LLMs for tasks that were subcategories of each usage type. Every participant could select multiple subcategories. The total number next to each title shows the number of participants who indicated using LLMs for the broad usage type.}
    \label{figure:usage-break}
\end{figure}

\section{Results}
\label{section:results}

\subsection{RQ1: What are the different ways researchers use LLMs in their research process today?}
\label{sec:rq1}

We asked participants to mark how frequently they used LLMs for each of six broad categories of research tasks.
When considering their answers across all six categories, we find that LLMs are now a common tool for researchers, with a total of \textbf{80.88\% (660 out of 816) of respondents adopting the use of LLMs somewhere in their research process}.
However, most usage today is still concentrated around tasks that have some connection to manuscript preparation, from ideation and framing of arguments, to editing and writing of text, to gathering information for literature review.
Still, some participants found many uses for LLMs; one prolific user reported using LLMs for \textit{``paper writing, analysis of data, results visualization, setting up new pipelines, `someone' to talk to for methods development and when I’m stuck - I have found myself doing all of the same work but getting through my todo list much faster.''}

Figure~\ref{figure:usage-frequency} shows the frequency of usage across all the respondents of our survey for our six high-level categories of LLM usage sorted by decreasing usage from left to right. As can be seen, there are major differences in usage across categories.
49\% and 45\% of our participants use LLMs for Information Seeking and Editing at least occasionally, respectively, while for categories related to data, such as Data Cleaning \& Analysis and Data Generation, most respondents (69\% and 73\%, respectively) stated they never used LLMs for these tasks.

We also see differences in usage within each of these categories when we break down the ways LLMs can be used.
We tally the number of people who selected if they have ever used LLMs to perform various tasks within each of the top-level categories and present the counts in  Figure~\ref{figure:usage-break}. The total referenced in each category is the number of respondents who stated they used LLMs for any of the tasks in the category, as well as an open-ended `Other' option. Respondents were able to select multiple options within each category.
More than a quarter of all respondents used LLMs for every one of the tasks under our Information Seeking category.
However, by far the most frequent usage of LLMs is for rewriting text to fix grammar or awkward phrasings, as used by almost half of all respondents, under the Editing category.
For the Data-related categories of usage, we see more usage in tasks relating to analysis and less usage in tasks related to simulation and synthetic data generation.

\begin{figure}
    \centering
    \includegraphics[width=\textwidth]{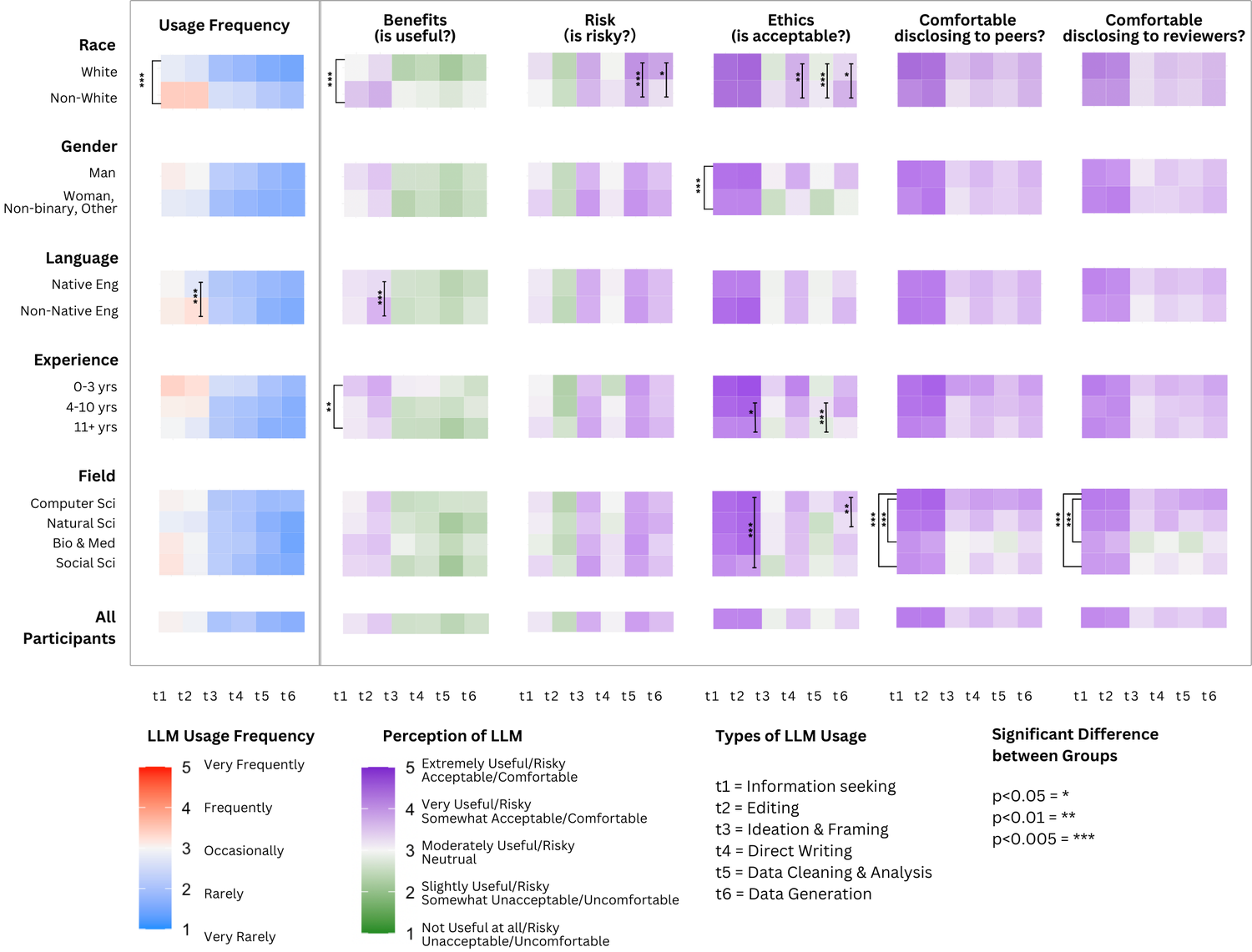}
    \caption{Overview of our survey results ($N$ is shown in Table \ref{tab:data-overview}), broken apart by demographic characteristics. Each heat map square represents the \textbf{average} rating of this demographic group on the usage frequency or perception for the particular type of LLM usage. The stars (***) are the significance levels of the differences indicated by p values from regression results. The brackets on the left indicate this difference is significant across all types of usage whereas the lines between squares indicate this difference is only significant for certain types of usage.}
    \label{figure:results-overview}
\end{figure}

\subsection{RQ2: How does the background of a researcher relate to the way they use LLMs?}
\label{sec:rq2}

Our second research question asks how usage varies according to the demographics, background, and other contextual factors related to the researcher.
In Figure~\ref{figure:results-overview}, the first column of heatmaps shows LLM usage frequency for each high-level category of LLM usage broken down by the five background characteristics that we surveyed.

Across all the LLM usage categories, we find that \textbf{researchers' racial identity significantly influenced the usage of LLMs}, with Non-White researchers ($\mu$ = 2.68, $\sigma$ = 1.73) reporting more frequent usage of LLMs than White researchers ($\mu$ = 2.06, $\sigma$ = 1.52; $Estimate$ = 0.616, $p$ < .0001).
Finally, we note that for the specific category of using LLMs for editing, we see significantly greater usage by NNES (Non-native English-speaking) researchers (\begin{math}Estimate=0.5069, p<.0001\end{math}), though this difference was not found for other categories of LLM usage, including using LLMs for direct writing.

\subsection{RQ3: What are researchers’ perceptions of LLMs for usage in different parts of the research process?}
\label{sec:rq3}

We ask participants their perceptions of the benefits and risks of LLM usage, how acceptable they perceive the use LLMs to be (i.e., ethics), and their comfort with disclosing LLM usage to peers and reviewers for each category of LLM usage using Likert rating questions.
We also asked participants to elaborate  on their answers via open-ended free-response questions.
In the final row of columns 2--6 of Figure~\ref{figure:results-overview}, we present the average Likert rating given by participants to different questions of perceptions broken down by category of LLM usage.

\subsubsection{Perceptions of LLM \textbf{benefits}}

Overall, as seen in Figure~\ref{figure:results-overview}, participants found greater benefits from LLM usage categories of Information Seeking ($\mu = 3.2$) and Editing ($\mu = 3.4$) compared to the remaining four categories ($\mu \leq 2.6$).
When we asked participants to elaborate on what specific benefits and usefulness they found, our analysis unveiled the following themes: language equity, other equity, efficiency, routine task assistance, search, literature review, editing, overcoming writer's block, broadening perspectives, programming, and brainstorming.
Most frequently mentioned were editing, literature review, efficiency, and language equity.
All of the themes are listed, along with descriptions and example quotations, in Table \ref{table:qual-themes-65} in Appendix \ref{appendix:qual-themes}.

Respondents stated they preferred using LLMs for smaller and routine tasks such as editing (\textit{``mostly rephrasing, rewriting, condensation, bulleting''}) and programming (\textit{``basically code using CoPilot + GPT now, often for research glue code.''}).
These low-level uses were especially highlighted in the context of researchers facing systemic barriers, such as non-native English speakers, junior scholars, and researchers without much programming experience (\textit{``I am not a native English speaker, so LLMs help me with the language barrier.''}).
Overall, \textbf{equity was a large theme in respondents' discussions of the benefits of LLMs} (\textit{``For honest researchers in resource-constrained developing countries, with little to
no research funding, availability and use of LLMs is a game-changer leveling the playing field with other researchers in more fortunate climes.''}).

Higher-level tasks were discussed by some users. For example, brainstorming  (\textit{``LLMs are a great tool to help you create hypotheses, as a way to brainstorm, where there really are no wrong ideas and therefore you cannot suffer with any potential misleading information, as you are expected to have domain expertise anyway.''}) and overcoming writer's block (\textit{``The major benefit of using LLMs comes in action when we are stuck, for example
not knowing the exact word or term, or not finding the answer to some of the ideas about how it can be used or how it can be applied.''}). At the same time, such mentions were usually followed by caveats, such as:  \textit{``You have to check everything it tells you, but it can be a useful starting place''}.

\subsubsection{Perceptions of LLM \textbf{risks} and \textbf{ethical concerns}}

While LLMs have risks related to the quality of research conducted, researchers could also have ethical concerns separate from issues of quality.
In order to distinguish risks from ethics, we first asked participants to rate on a Likert scale their perception of risks (1: not risky at all -- 5: extremely risky), given known issues with LLMs today.
Next, we asked their perception of the acceptability (1: Unacceptable -- 5: Acceptable) of using LLMs given a future where LLMs can prevent hallucinations and can always attribute any copyrighted text (if generated) to the original sources (i.e., their perception of the ethics of using LLMs).

Overall, as seen in Figure~\ref{figure:results-overview}, we find that researchers perceive using LLMs for Editing as not risky ($\mu = 2.5$), Direct Writing as moderately risky ($\mu = 3$), and the remaining categories as very to extremely risky ($\mu \geq 3.2$).
In contrast when it came to ethics, researchers find Ideation \& Framing ($\mu = 2.9$) and Data Cleaning \& Analysis ($\mu = 2.96$) to be more on the unacceptable side, while the remaining categories were found to be more acceptable ($\mu \geq 3.4$).

We asked one free-response question which gave participants the option to elaborate on risks and ethics together.
Our thematic analysis of responses unveiled the following themes: hallucination and misinformation, inaccuracy, biases, lack of disclosure, plagiarism, disrespecting authorship, fabrication, decreasing creativity, pollution of the research ecosystem, decreasing diligence, and deskilling.
Most frequently mentioned were hallucination and misinformation, plagiarism, fabrication, and decreasing diligence.
All of the themes are listed, along with descriptions and example quotations, in Table \ref{table:qual-themes-66} in Appendix \ref{appendix:qual-themes}.

Qualitatively, based on the optional free-text responses, we found that our respondents have \textbf{strong opinions about the risks and ethics of LLMs for research}.
They frequently use strong language to describe their positions (\textit{``LLMs are tools for automated plagiarism and data fabrication that pose an existential threat to the network of trust essential for the integrity of academic work and the proper attribution of credit''}).
While many point to specific risks like data fabrication (\textit{``the risk of reporting `results' based on synthetic data without actually having conducted any experiment''}) and plagiarism (\textit{``blind trust in a system that is hard to understand which could lead to accidental plagiarism''}), others draw attention to higher level concerns that could affect all of academic research, such as pollution of the research ecosystem with low-quality work (\textit{``We need better judgment, slower science, and more thoughtful and ambitious work right now, not the opposite. Otherwise, we risk ridding science of its most special attributes just to crank
out more papers.''}).

Many respondents worried about future generations of researchers whose skills, diligence, and creativity might be impacted by over-reliance on LLMs (\textit{``The main general risk is to flatten on `average', which is the worst thing that may happen for a researcher (and it is already happening for arts such music, since this would block innovation.''}).
Researchers also worried about exacerbating existing problems, such as overwhelming numbers of papers needing review (\textit{``I fear for a deluge of AI-`assisted' (in the best case) papers that read somewhat fluently but are shallow, unoriginal, uninteresting, wrong in the details. This will overwhelm the peer review system''}).

\subsubsection{Comfort with disclosure}

Finally, we asked participants to rate on a Likert scale their comfort with disclosing the use of LLMs along different LLM usage categories (1: uncomfortable -- 5: comfortable). Overall, as seen in Figure~\ref{figure:results-overview}, participants were \textbf{comfortable with disclosing to both peers and reviewers across all LLM usage categories} (disclose to peers: $\mu \geq 3.4$, disclose to reviewers: $\mu \geq 3.2$).

When it came to our qualitative results, participants' opinions on the
disclosure of LLM usage and proper attribution were discussed across multiple survey questions and our extracted themes, though respondents particularly discussed disclosure in response to the question about risks and ethical considerations.
One respondent mentioned \textit{``academic shame''} as a reason researchers might not disclose LLM usage.
Other respondents highlighted the costs to the research community of not disclosing LLM usage: \textit{``[If researchers don't disclose using LLM-generated text], I fear that researchers can get lazy, and we start having a lot of `repeated text' in articles... and eventually researchers may just ask LLMs to generate the whole paper.''}
Some respondents listed this as their main concern with model-assisted research, and as long as LLM usage was disclosed, they found that usage acceptable: \textit{``The same sort of disclosure of use [as with human assistance] should be sufficient. The same responsibility for the integrity of work applies whether part of the effort was provided by a human assistant or an LLM.''}
Finally, one user called for better processes to support disclosure: \textit{``...universities have totally different policies. It would be good if there was a generic system of how to indicate that editing or drafting tools were used.''}

\subsection{RQ4: How does the way a researcher uses LLMs relate to their perceptions?}
\label{sec:rq4}

\begin{figure}
    \centering
    \includegraphics[width=\textwidth]{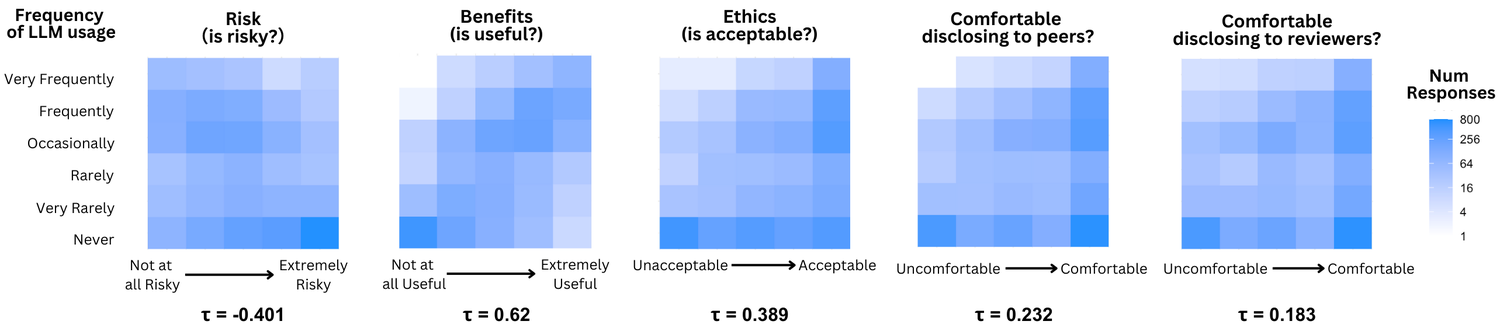}
    \caption{Overview of the relation between usage frequency and perceptions of LLM. Each heatmap represents one type of perception, and each cell represents the number of responses (log scaled) that fall under this level of frequency of perception. The Kendall's tau coefficient on the bottom indicates how strong the correlation is between the usage frequency and the perception of that usage. All perceptions are significantly correlated with usage ($p$ < .0001). Tests performed using \texttt{cor.test} in \texttt{R} and corrected with \texttt{p.adjust}.}
    \label{figure:frequency-perception}
\end{figure}

We find that people's perceptions of risks, benefits, and ethics and their willingness to disclose usage of LLMs to the community all had a significant impact on their stated usage of LLMs ($p < .0001$).
Figure \ref{figure:frequency-perception} presents heatmaps showing the number of responses (log scaled) for each level of usage frequency and perception level. These values are summed across all six categories of LLM usage, i.e., each respondent is represented six times in each heatmap if they answered all questions about usage and perceptions. In addition, Table~\ref{tau-correlation} in Appendix~\ref{appendix:survey-stats} presents the correlation between perceptions and frequency of LLM usage broken down by type of LLM usage.

Overall, as expected, we see that \textbf{greater perceived risks and greater perceived ethical concerns are associated with lower usage, and greater perceived benefits are associated with higher usage}.
However, \textbf{perceived benefits has the strongest correlation to usage} (0.62, $p$ < .0001). As shown in Figure \ref{figure:frequency-perception}'s heatmaps, some who find few risks or ethical concerns with certain categories of usage still choose to use LLMs for them infrequently or not at all, potentially because they find little benefit.
Indeed, when looking at Table~\ref{tau-correlation} in the Appendix, we can see overall a weaker relationship between risk and frequency (-0.401, $p$ < .0001) and ethics and frequency (0.389, $p$ < .0001) compared to benefits.
While this is also true for every category of LLM usage, in certain categories, risks and ethics have a stronger relationship with (non-)usage, such as Direct Writing.

We find a weaker but positive relationship between comfort with disclosure and frequency of LLM usage. Looking at the heat maps, we see that the majority of researchers feel comfortable disclosing LLM usage to both their peers and reviewers regardless of their usage level, with the exception of those who never used LLMs for a given category, where a sizeable portion expressed discomfort.
This suggests greater discomfort with disclosure, or more broadly lack of social norms around usage, could be one reason why some researchers do not wish to use LLMs for one or more categories of research.

\subsection{RQ5: How does the background of a researcher relate to their perceptions?}
\label{sec:rq5}

\input{tables/pvalues}

In Figure~\ref{figure:results-overview}, columns 2--6 show heat maps related to respondents' perceptions of using LLMs across all six usage categories and broken down by respondents' background.
Table \ref{tab:pvalues} also presents results of significance tests for differences.

\subsubsection{Differences in perceptions of LLM \textbf{benefits}}

We find that across all LLM usage categories, a researcher's \textbf{race and years of experience have a significant effect on their perception of the benefits of LLMs}.
Non-White researchers ($\mu = 3.14$, $\sigma = 1.31$) perceive more benefits in using LLMs for research than White researchers ($\mu = 2.67$, $\sigma = 1.35$), while junior researchers ($\mu = 3.20$, $\sigma = 1.29$) with 0--3 years of experience perceive more benefits than senior researchers ($\mu = 2.76$, $\sigma = 1.38$) with 11+ years of experience.
Given our results regarding usage frequency across backgrounds (RQ2) and correlation between perceptions and usage (RQ3), this suggests that perceived utility is a primary driver of greater usage for non-White researchers and junior researchers.
Similarly, we find that NNES researchers perceive greater benefits in using LLMs for Editing than native English-speaking researchers (\begin{math}Estimate=0.4187, p=0.0004\end{math}), in addition to actually using LLMs for Editing significantly more as well.

In general, the groups who reported perceiving more benefits are some of the groups who are traditionally less advantaged in the research community: non-White researchers, non-native English speakers, and researchers with the least experience.
These findings support arguments that LLM usage can potentially play a role in improving research equity, echoing our qualitative results.

\subsubsection{Differences in perceptions of LLM \textbf{risks}}

Overall, there were few significant differences in how people of different backgrounds perceived the risks of LLMs across all the LLM usage categories.
One exception was between White and non-White researchers, where non-White researchers perceive fewer risks in using LLMs for Data Cleaning \& Analysis (\begin{math}Estimate=-0.573, p<.0001\end{math}), and Data Generation (\begin{math}Estimate=-0.274, p=0.0352\end{math}) than White researchers.
This suggests that a perception of heightened risk may be what depresses White respondents' usage of LLMs for these categories in addition to lower perceived benefits.

\subsubsection{Differences in perceptions of the \textbf{ethics} of using LLMs}

We find that across all LLM usage categories, a researcher's \textbf{gender has a significant effect on their perception of the ethics of using LLMs}.
Overall, researchers who identify as women, non-binary, and other genders ($\mu = 3.24, \sigma = 1.58$) perceive LLM usage in research as less acceptable than those who identify as men ($\mu = 3.6, \sigma = 1.47$) as shown in Table \ref{tab:pvalues}.
This suggests that ethical concerns are a major driver for lower LLM usage for women and non-binary researchers compared to men, though the differences there in terms of usage were not significant.

We also find differences in ethical concerns across backgrounds according to specific LLM usage categories.
Similarly to perceptions of risk, we find that non-White researchers perceive fewer ethical concerns compared to White researchers for the two LLM usage categories related to data (Data Cleaning \& Analysis: \begin{math}Estimate=0.4059, p=0.0027\end{math}, Data Generation: \begin{math}Estimate=0.3066, p=0.0230\end{math}) as well as for Direct Writing (\begin{math}Estimate=0.3391, p=0.0098\end{math}).
This suggests that differences in ethical concerns along with perceptions of risk and benefits may contribute to differences in White and non-White researchers' LLM usage.

We also see differences in ethical concerns between more junior (4--10 years of experience) and more senior researchers (11+ years of experience) for the categories of Editing (\begin{math}Estimate=0.3385, p=0.0215\end{math}) and Data Cleaning \& Analysis (\begin{math}Estimate=0.4578, p=0.0013\end{math}), where senior researchers have greater ethical concerns.
This suggests that in addition to differences in perceptions of benefits, differences in ethical concerns may drive what differences there are in LLM usage between more junior and senior researchers, though the usage differences we find are not significant.

Finally, we find significant differences in perception of ethics across research fields for some LLM categories of usage.
Computer science researchers on the whole have fewer ethical concerns than researchers in other fields.
In particular, computer science researchers perceive using LLMs for Editing as more acceptable than social science \& humanities researchers (\begin{math}Estimate=0.5706, p=0.0008\end{math}), and they also perceive Data Generation as more acceptable than natural science \& engineering researchers (\begin{math}Estimate=0.5279, p=0.0075\end{math}).
Despite not seeing greater usage of LLMs by computer scientists than researchers in other fields, it is possible that usage may yet be more normalized in computer science due to LLMs being a major topic of active research.

\subsubsection{Differences in comfort with disclosure}
We find that a researcher's \textbf{field of research has a significant effect on their comfort with disclosing usage of LLMs to peers and reviewers}.
In particular, computer scientists (To peers: $\mu = 4.07, \sigma = 1.39$; To reviewers: $\mu = 3.91, \sigma = 1.47$) are more comfortable disclosing LLM usage than social science \& humanities (To peers: $\mu = 3.52, \sigma = 1.57$; To reviewers: $\mu = 3.42, \sigma = 1.58$) or biology \& medicine researchers (To peers: $\mu = 3.37, \sigma = 1.61$; To reviewers: $\mu = 3.21, \sigma = 1.65$).
This finding echoes our findings related to ethical concerns, which taken together further suggests greater acceptance in the computer science research community around the use of LLMs compared to other fields.
Interestingly, we see no significant differences across other aspects of researcher background with regards to comfort with disclosure; indeed as mentioned earlier, people are mostly comfortable with disclosure to both peers and reviewers across all LLM categories of usage.

\subsection{RQ6: How does the source of the LLM affect researchers’ perception and usage?}
\label{sec:rq6}

Participants were split on whether the source of an LLM, (i.e., non-profit versus pro-profit entities), impacted their perception of benefits and risks.
54.81\% of participants (359) reported that their perception would change depending on the source of LLM while 45.19\% (296) reported no difference.
We also asked participants to optionally elaborate on their selection in an open-ended question.
From manual coding of responses, we found that \textbf{59.07\% (228) of respondents expressed a preference for LLMs from open source/non-profit entities,\footnote{In the free response, some participants use open source and non-profit interchangeably. Thus, for the seek of labeling, we create a higher-level label of open source/non-profit to capture that opinion.} while only 2.85\% (11) stated they preferred LLMs from for-profit corporations}, and 38.08\% (147) did not express a preference either way in their free-response answer.

We also qualitatively coded responses for themes regarding why people preferred non-profit or for-profit entities behind LLMs; full results are shown in Table \ref{table:qual-themes-64} in Appendix \ref{appendix:qual-themes}.
The top reasons participants preferred non-profit include the \textit{incentives} of the organization, the \textit{transparency} of the model, and \textit{ethical} considerations for LLMs.
These participants were skeptical of the incentives of commercial corporations, and worried that they would ``\textit{exploit user input, manipulate LLM outputs for financial gain}.'' They also expressed concerns about monopolization, injecting bias to maximize profits, and other ethical issues. They favored non-profit entities because of the transparency in open-source models, increasing accountability and users' trust.

From the few participants who actually favored LLMs from commercial corporations, they stated as a rationale that they believed those models were of higher quality due to the resources available to companies and their responsibility towards supporting customer issues.
For participants who perceived no difference in the source of the LLMs, some held a neutral attitude that ``\textit{the technology is the same}'' no matter which organization provided it.
Some expressed that they cared more about the quality of the model, and they would use the model with the best quality regardless of its source.
Finally, other participants questioned the boundary between for-profit and non-profit entities: ``\textit{as we have seen with OpenAI, non-profits can easily become commercial}.''

For some respondents, whether the model itself is open-source was more important than whether the LLM was created by a non-profit or for-profit entity.
Other respondents expressed skepticism with the open-source label: ``\textit{No LLM is really open source. Most of them owe their existence to big commercial corporations, and even if they share the weights, we don't really know all the details about the training data. They are essentially black boxes}.''

\section{Discussion}

Results from our survey showed widespread adoption of LLMs in the research community and provided detailed insights into the different ways researchers leverage LLMs in their workflows. We found that while LLMs offer the potential for enhancing equity and productivity, particularly benefiting those less advantaged in the research community, they also raise concerns about research integrity, quality, and potential homogenization of scholarly output. The varying  levels of LLM usage and comfort levels with disclosure across disciplines and career stages highlight the ongoing negotiation of new social norms in academia.

\subsection{Deep and Pervasive Integration of LLMs in Research}
Our work revealed that many researchers have already found benefits in incorporating widely available LLM-based tools into their current workflow, from literature review to data analysis to writing assistance.
This confirms and expands upon prior assumptions about the prevalence of LLM usage in academia~\cite{morris2023scientists, Gruda_2024, bail2024can, koller2023we, Kobak_2024}. While we did not set out to explore how researchers describe their relations with LLM-based research support tools, many in the free-form responses explicitly describe these tools with varying levels of autonomy and agency. More specifically, the ways participants described LLMs ranged from direct manipulation~\cite{shneiderman1982future} (``\emph{just another tool in the toolbox}'') to data sources (``\emph{a custom Wikipedia page}'') to human-AI teaming~\cite{10.1145/3432945} (``\emph{a useful research collaborator or assistant},'') to fully autonomous agents~\cite{10.1145/267505.267514} (``\emph{an end-to-end AI researcher}''), pointing to a wide design space of future LLM-based research support tools and user interfaces \cite{10.1145/3613904.3642697}. As researchers across AI and HCI domains continue to devote resources to developing new tooling \cite{gero2022sparks,10.1145/3613904.3642698,wang-etal-2024-scimon,lo2023semantic,fok2023qlarify,lee2024paperweaver,kang2023synergi,kim2023metaphorian,long2023tweetorial, Gruda_2024,sun2024metawriter,d2024marg,hsu2024chimellmassistedhierarchicalorganization},
we may see increasing benefits in adopting LLM-based research support tools and a potential paradigm shift in how research is conducted in the future.

\subsection{``\textit{A Game-changer Leveling the Field}'': Equity Benefits of LLMs}
An unexpected finding was the frequent mention of \emph{equity} as one of the main benefits of using LLMs for research. Most frequently, non-native English speakers described how LLMs allowed them to ``\emph{level the playing field}'' by cutting down ``\emph{tedious and time-consuming editing tasks}'' to ``\emph{more freely and precisely express ideas in another language [English]}.''
Our quantitative findings in Section~\ref{sec:rq2} also reveal that groups traditionally disadvantaged in research \cite{linxen2021weird} (non-White, junior, non-native English speakers) find LLMs more beneficial and, in some cases, use them more frequently.
Additionally, equity was mentioned in other contexts such as enabling researchers without technical programming training to generate code for data cleaning, improving understanding of technical papers and papers from less familiar fields, or replacing high-cost editing and proofreading services.
For example, one neurodivergent participant pointed to how writing with LLMs allowed them to write more confidently and productively.
This impact suggests that LLMs are beneficial for researchers from various marginalized backgrounds, helping them overcome systemic barriers --- including ``knowledge abysses'' and neo-colonial dynamics in research~\cite{Goyes2023RichSP} --- and gain deserved visibility in the global research community.

However, ethical concerns surrounding LLMs could potentially hinder their broader adoption, particularly among certain demographic groups.
From the results in Section~\ref{sec:rq5}, we find that women and non-binary participants in our study expressed greater ethical concerns about LLM use and tend to use them less frequently (though not statistically significant).
This pattern is particularly noteworthy given that women and non-binary researchers are traditionally disadvantaged groups in academia.
If these researchers abstain from using LLMs due to ethical concerns and thus lose out on potential benefits or are shut out of potential collaborations, this could further exacerbate inequities.
Future work should examine more deeply the specific ethical and other concerns expressed by traditionally disadvantaged groups and strive to address these concerns in future development of LLM-based research tools.
For instance, given prior research showing women scientists often struggle to receive appropriate credit for their work~\cite{ni2021gendered,ross2022women}, women may be wary of using LLMs or attributing some aspect of their work to them.
In our survey, we did not see definitive evidence of gender playing a role in LLM usage or comfort with disclosure, though future work could go beyond self-report data and personal perceptions to examine observational data or impacts of LLM attribution.

\subsection{Productivity vs. Research Integrity}
Despite claimed benefits in research productivity and equity, our survey also reveals significant concerns about the risks associated with LLM use in research.
Hallucinations and misinformation were among the most frequently mentioned risks.
Some participants used strong language, such as plagiarism and data fabrication, and others expressed fundamental concerns about the impact of LLMs on future generations of researchers, potentially affecting their skills, diligence, and creativity, which might result in the proliferation of low-quality research~\citep{bail2024can}.
These concerns are reflected by academic organizations and funding agencies; for example, the NIH has cautioned against using LLMs for applications or reviews because it sacrifices the ``\emph{originality of thought}'' and leads to homogenization of ideas or even research misconduct, subject to the severe penalty~\cite{Lauer_Constant_Wernimont_2023}.

Interestingly, our survey showed that while disadvantaged groups are more likely to discuss \textit{benefits} of LLMs, as discussed earlier, we observe few significant differences in perceptions of \textit{risks} across different backgrounds.
This suggests that the perception of risks appears to be more uniformly distributed across demographics.
For example, the problem of hallucinations in LLM outputs appears to be an equally significant concern for all researchers.
This shared understanding indicates a collective awareness of LLMs' limitations and suggests the possibility of developing uniform standards for LLM use that can be broadly agreed upon, irrespective of people's demographic characteristics.

\subsection{Emerging Norms and Guidelines}
Scholarly venues and research funding agencies have begun discussing the ban of LLMs in research or peer reviews~\citep{Grove_2023, Kaiser_2023}.
However, the ban on LLMs in research has been criticized as undesirable given its potential benefits and was unenforceable due to undisclosed use of LLMs~\citep{hosseini2023ethics}.
The more realistic approach might be establishing guidelines about the ethical use of LLMs in scientific research.
Several articles have proposed such guidelines.
~\citet{watkins2023guidance} created a peer-reviewers' checklist for the use of LLM agents.
\citet{hosseini2023ethics} advocates for transparent disclosure of LLM use, including specific citation details, and inclusion of LLM interactions as supplementary material for reproducibility.
\citet{koller2023we} elicits best practices for using LLMs in various knowledge work scenarios such as viewing LLMs as aids rather than collaborators and engaging in post-processing of LLM-generated content to maintain a sense of ownership and control.
In the context of software development research, \citet{Sallou_Durieux_Panichella_2024} outlines guidelines including providing reproducibility metadata and assessing output variability by performing multiple runs.
These growing norms aim to promote transparency, reproducibility, and thoughtful integration of LLM as research tools while maintaining high standards of scientific rigor.

\subsection{Weighing the Impacts of Mandatory LLM Disclosure in Academia}
One of the most common approaches to establishing norms on LLMs in research is to increase transparency.
Many conferences, publishers, and funding agencies have started to implement guidelines and restrictions for LLM use, and/or require researchers to disclose how LLM-based tools were used during research or grant writing ~\cite{acl,acm,aaai,ieee,2023WritingTR,nih,Kwong2024TheLB}.
Our survey findings generally align with this trend, revealing that participants are broadly comfortable with disclosing LLM use to both peers and reviewers across all categories of usage.
While we do not observe major differences in usage across fields, our study uncovered nuanced variations in comfort levels.
Computer scientists reported significantly higher comfort with disclosure compared to researchers in social sciences \& humanities or biology \& medicine.
This disparity reflects the evolving and unsettled nature of LLM use in research across different academic fields. While some disciplines may be moving towards accepting LLMs as standard research tools, others are still grappling with how to integrate these technologies into their established practices. The mention of ``\emph{academic shame}'' by one respondent as a potential reason for non-disclosure highlights the ongoing stigmatization of LLM use in some research communities. The lack of standardized policies across institutions further complicates this landscape, creating additional burdens for individual researchers who must navigate varying expectations and norms.
This suggests that the academic community is still collectively negotiating the norms of acceptable LLM use, and continued discussion is crucial to fully leverage new technologies to improve science while avoiding potential pitfalls.

\subsection{Open-Source vs. Closed-Source LLMs}
The debate between open-source and commercial LLMs adds another layer of complexity to their use in research. Our survey revealed that while commercial models are perceived to offer higher quality and better user support, they also raise significant concerns about transparency, reproducibility, and more fundamentally, the misalignment of incentives between commercial entities and the research community.
Many researchers, particularly those lacking technical expertise or computational resources, predominantly rely on commercial closed models~\citep{Wulff2024open, Toma2023generative, Hussain2023open}.
However, this reliance comes with risks.
For example, companies deprecating earlier versions of their hosted models\footnote{``As we launch safer and more capable models, we regularly retire older models.
Software relying on OpenAI models may need occasional updates to keep working.'' \url{https://platform.openai.com/docs/deprecations}} could potentially render many thousands of published research papers significantly less reproducible~\cite{Kapoor_Narayanan_2023}.
This risk aligns with a growing consensus in the research community that open-source LLMs offer significant advantages for scientific work.
Many researchers argue that the use of open-source models enhances the validity and integrity of research by allowing for greater scrutiny of research data and output~\citep{Sallou_Durieux_Panichella_2024, Toma2023generative}.
This stands in strong contrast to closed-source models that can be obscured behind opaque APIs; one notable example being the recent release of GPT-o1 model that hides parts of its generations from users.\footnote{``Each completion has an upper limit on the maximum number of output tokens—this includes both the invisible reasoning tokens and the visible completion tokens.''
\url{https://platform.openai.com/docs/guides/reasoning}}
These concerns should be carefully examined in high-stakes research areas such as medicine, bioengineering, law, and public policy, which can have direct, real-world impacts~\citep{Toma2023generative}.
In these fields, reproducibility and transparency are paramount, as they ensure the reliability of findings and provide necessary justification for decisions that affect people's lives.

\subsection{``\emph{Not Me, But Them}'': Self vs. Peer Perceptions Discrepancy}
Our study reveals an interesting discrepancy in how researchers perceive their own use of LLMs versus that of other scholars.
Researchers appear to trust their own judgment in using LLMs for limited, low-risk tasks, while expressing concerns about more extensive or inappropriate use by others.
This mismatch between stated fears and reported personal usage is exemplified by responses such as, ``\emph{I'm afraid other people will use models to write papers but I only report using it for editing}.''
This phenomenon aligns with established concepts in social psychology and media studies: the ``above-average effect''~\citep{kruglanski1990classic}, where individuals tend to view themselves as superior to their peers, and the ``third-person effect''~\citep{Davison_1983}, which describes how people often believe they are less susceptible to media influence than others. This perception might facilitate the establishment of community-wide norms for LLM use as researchers may be more receptive to stricter guidelines, believing that while they have already implemented self-regulation, others in the community require more oversight.

\section{Conclusion}

In this study, we ran a large-scale survey of diverse groups of researchers, soliciting information about their usage of LLMs for their work as well as their perceptions of LLM usage by other researchers.
Our study provides lessons about the changing social norms across different academic disciplines and demographic groups, and we found significant differences in responses between these groups.
Researchers reported widespread adoption of LLMs in their work, with 80.9\% of surveyed researchers using them, primarily for information seeking and editing tasks.
Non-White researchers, junior scholars, and non-native English speakers reported using LLMs more frequently and perceiving greater benefits, suggesting LLMs may help level the academic playing field.
However, researchers who identified as women, non-binary, or other genders expressed greater ethical concerns about LLM usage.
While perceived benefits strongly correlate with usage frequency, researchers also acknowledge significant risks, including hallucinations, misinformation, plagiarism, and potential long-term impacts on research quality and creativity.
The majority of respondents prefer LLMs from open-source and non-profit entities, citing better incentive structures, transparency, and ethical considerations.

Collectively, our work suggests that the research community is at a critical juncture, grappling with upholding fundamental values of originality, rigor, and ethical conduct in academia.
As \citet{Brinkmann2023machine} terms it, we are currently shaping ``Machine Culture,'' where technologies like LLMs serve as cultural mediators and generators, capable of transforming cultural evolutionary processes.
As we continue to progress on both the capability of LLMs and LLM-based research support tools to provide greater benefits, the integration of LLMs into research practices is also likely to continue to increase in both depth and breadth.
This represents a significant shift in this cultural evolution, with the potential to pose risks to research integrity and the development of critical thinking skills among researchers or, if carefully managed, contribute to a more productive, diverse and inclusive global research community.
Our findings underscore the need to better understand the implications of LLM use across various research contexts.
We should explore not only the technical aspects of LLM integration but also its sociological, ethical, and epistemological impacts on different disciplines and researchers' demographic backgrounds.
We call for studies that examine the long-term effects of LLM use on research quality, creativity, and the development of research skills as well as investigations into the potential of LLMs to increase fairness and representation in academia.

\begin{acks}
The authors would like to thank our many anonymous survey respondents who provided long and thoughtful opinions and insights in the optional free-text survey questions. These responses enabled our qualitative analysis and showcased that many researchers are actively contemplating and engaging in conversations around the use of LLMs as a research support tool today.
\end{acks}

\bibliographystyle{ACM-Reference-Format}
\bibliography{REFERENCES}

\newpage
\appendix
\section{Survey Statistics}
\label{appendix:survey-stats}

\input{tables/Data_overview}

\input{tables/tau_correlation}

\input{tables/chisq_goodness_fit_demographics}

\section{Quantitative Method}
\textbf{Creating the Field Demographics Categories}\\
\label{appendix:creating-field}
We collected the research fields that our participants studied in the form of free responses. Out of the 816 responses, 644 (79\%) responses included field information. We classified 635 free responses into four field categories: computer science, social science \& humanities, natural science \& engineering, and biology \& medicine, and 9 responses were classified as other and excluded from the analysis. Computer science group had 257 participants, and also included interdisciplinary fields with computer science, such as education technology, except biotechnology. Social science \& humanities had 152 participants, and included psychology, behavioral science, education, sociology, and more. Natural science \& engineering had 132 participants, and included math, chemistry, physics, environmental science, electrical engineering, and more. Biology \& medicine had 94 responses, and included cognitive science, bioinformatics, biotechnology, public health, neuroscience, and more.

\section{Qualitative Themes}
\label{appendix:qual-themes}

\input{tables/qual_themes_64}
\input{tables/qual_themes_65}
\input{tables/qual_themes_66}

\end{document}

%% file: COMMANDS.tex
\definecolor{table-row-color}{HTML}{eaeaea}

\definecolor{maria-color}{HTML}{7881F2}

\definecolor{kyle-color}{HTML}{0fcb8c}

\definecolor{simona-color}{HTML}{228B22}

\definecolor{inyoung-color}{HTML}{cb6ce6}

\definecolor{todo-color}{HTML}{ff3232}

\definecolor{joseph-color}{HTML}{FF69B4}

\definecolor{amy-color}{HTML}{ff0000}

\newcommand{\platform}{{Semantic Scholar}}

%% file: tables/chi_results_subfigures.tex
\begin{table}[h!]
    \centering
    \footnotesize
    \begin{subfigure}{0.48\textwidth}
        \centering
        \begin{tabular}{ccc}
            \toprule
            & Man & \makecell[tl]{\footnotesize Woman, Non-\\ \footnotesize Binary, Other} \\
            \midrule
            White & 243 & 74 \\
            Non-White & 168 & 34 \\
            \bottomrule
        \end{tabular}
        \caption*{$p = 0.7578$}
        \caption{Race and Gender}
    \end{subfigure}
    \hfill
    \begin{subfigure}{0.48\textwidth}
        \centering
         \footnotesize
        \begin{tabular}{lccc}
            \toprule
            & 0-3 & 4-10 & 11+ \\
            \midrule
            White & 27 & 96 & 198 \\
            Non-White & 31 & 81 & 94 \\
            \bottomrule
        \end{tabular}
        \caption*{$p = \textbf{0.0101}$}
        \caption{Race and Years of Experience}
    \end{subfigure}

    \vskip\baselineskip

    \begin{subfigure}{0.48\textwidth}
        \centering
         \footnotesize
        \begin{tabular}{lcccc}
            \toprule
            & CS & Bio & Nat.Sci & Soc.Sci \\
            \midrule
            Man & 194 & 65 & 113 & 110 \\
            \makecell[tl]{\footnotesize Woman, Non-\\ \footnotesize Binary, Other} &  \raisebox{-1ex}{48} & \raisebox{-1ex}{22} & \raisebox{-1ex}{12} &  \raisebox{-1ex}{40}  \\
            \bottomrule
        \end{tabular}
        \caption*{$p = \textbf{0.0008}$}
        \caption{Gender and Years of Experience}
    \end{subfigure}
    \hfill
    \begin{subfigure}{0.48\textwidth}
        \centering
         \footnotesize
        \begin{tabular}{lcccc}
            \toprule
            & CS & Bio & Nat.Sci & Soc.Sci \\
            \midrule
            0-3 & 37 & 6 & 11 & 12 \\
            4-10 & 102 & 30 & 39 & 35 \\
            11+ & 118 & 58 & 82 & 105 \\
            \bottomrule
        \end{tabular}
        \caption*{$p = \textbf{0.0030}$}
        \caption{Years of Experience and Field of Study}
    \end{subfigure}
    \caption{Contingency tables of participant counts for different demographic pairs. $p$-values from Chi-square tests of independence.
    }
\label{tab:chi_results_subfigs}
\end{table}

%% file: tables/pvalues.tex
\begin{table}[t]
\centering
\begin{tabular}{lcccccc}
\toprule & \makecell{Man-\\Woman, Non-binary, Other} & $p$ & \makecell{Non-White-\\White} & $p$ & \makecell{11+ years-\\0-3 years} & $p$\\
\midrule
Benefit (1-5) &  & & 0.420 & <.0001 & -0.4187 & 0.0004   \\
Ethics (1-5) & 0.351 & 0.0017 &&&&\\\hline
\toprule & Bio-CS & $p$ & CS-Soc.Sci & $p$ & &\\
\midrule
Disclosure-Peers (1-5) & -0.644 & 0.0001 & 0.513 & 0.0002 &&\\
Disclosure-Reviewers (1-5) & -0.622 & 0.0007 & 0.451 & 0.0043 &&\\
\bottomrule
\end{tabular}
\caption{Post-hoc tests for significant pairwise differences in LLM perception ratings between demographic groups (Gender, Race, and Year on the first row, and Field of Study on the second), across all LLM usage types. This table reports the rating differences between demographic levels and associated $p$-values from \texttt{emmeans}. For example, in the column for race and the row for Benefit, the result shows that Non-White researchers, on average, report 0.42 points higher ratings than White researchers on the perceived benefits of LLM usage. Only statistically significant results are included in the table.}
\label{tab:pvalues}
\end{table}

%% file: tables/Data_overview.tex
\begin{table}[h!]
\centering
\begin{tabular}{lcccccc}
\toprule
& \textbf{Gender} & \textbf{Race} & \textbf{Language} & \textbf{Years} & \textbf{Field} & \textbf{All}\\ 
\midrule
RQ2 \hspace{0.64em}Frequency & 3666 & 3162 & 3864 & 3864 & 3810 & 4896\\ 
RQ5.1 Risk & 2973 & 2597 & 3110 & 3110 & 3066 & 3410\\ 
RQ5.2 Benefits & 2315 & 2016 & 2440 & 2440 & 2403 & 2738\\ 
RQ5 3 Ethics & 3191     & 2744 & 3354 & 3354    & 3314 & 3665\\ 
RQ5.4 Disclosure to Peers & 3061 & 2631 & 3207 & 3207 & 3168 & 3493\\
RQ5.4 Disclosure to Reviewers & 3028 & 2637 & 3177 & 3177 & 3139 & 3450\\
\bottomrule
\end{tabular}
\caption{Number of answers to survey questions across all participants, broken out by demographic and question. Each question (row) had six sub-questions; participants did not have to answer all questions.}
\label{tab:data-overview}
\end{table}

%% file: tables/tau_correlation.tex
\begin{table}[h!]
\centering
\begin{tabular}{lccccc}
\toprule \textbf{Type of Usage} &	\textbf{Risk}	& \textbf{Benefit}	& \textbf{Ethics}	& \textbf{Disclosure to Peers} 	& \textbf{Disclosure to Reviewers} \\

\midrule
Information Seeking	& -0.303	& 0.513	& 0.261	& 0.127 (0.0005)	&0.078 (0.041)\\
Editing	& -0.276 & 	0.557 &	0.282 &	0.171 & 0.073 (0.041)\\
Ideation \& Framing	& -0.372 &	0.651 &	0.385 &	0.197 &	0.177 \\
Direct Writing &	-0.401 &	0.575 &	0.409 &	0.184&	0.136 (0.0004)\\
Data Cleaning \& Analysis &	-0.437 &	0.68 &	0.335 &	0.229  & 0.186 \\
Data Generation &	-0.45 & 0.665	& 0.383	& 0.262 & 0.245\\ \midrule
Overall Usage &	-0.401	& 0.62 & 0.389 &	0.232 &	0.183 \\

\bottomrule

\end{tabular}
\caption{Kendall's tau correlation between the frequency of LLM usage and the perception of that usage. Each cell includes the tau coefficient with the Holm-Bonferroni corrected $p$-value in parenthesis. Coefficients not followed by parenthesis all had $p$<.0001.}
\label{tau-correlation}
\end{table}

%% file: tables/chisq_goodness_fit_demographics.tex
\begin{table}[h]
\centering
\begin{tabular}{lccccc}
\toprule
& Gender & Race & Year & Language & Field \\
\midrule
Gender & / & 0.7578 & 1.0000 & 0.1433 & \textbf{0.0008} \\
Race & & / & \textbf{0.0101} & 1.0000 & 1.0000 \\
Year & & & / & 1.0000 & \textbf{0.0030} \\
Publication & & & & 0.8043 & 1.0000 \\
Language & & & & / & 1.0000 \\
Field & & & & & / \\
\bottomrule
\end{tabular}
\caption{$p$-values from Chisquare tests of independence between demographic groups. All $p$-values were corrected via Holm-Bonferroni. Significant values ($p < 0.05$) are in \textbf{bold} and indicate dependence between pairs of variables.}
\label{tab:chisq-demographic-p-values}
\end{table}

%% file: tables/qual_themes_64.tex
\begin{table}[H]
    \centering
    \footnotesize
    \begin{tabular}{@{}p{2cm}|p{4cm}p{8cm}@{}}
    \toprule
    \textbf{Theme} & \textbf{Description} & \textbf{Example} \\
    \midrule
    
    \textbf{Reproducibility} 
    & Whether research is reproducible/replicable/verifiable by other researchers 
    & \textit{``Because research should be reproducible, and this is only possible with the use of open-source LLMs.''} \newline \textit{``...an open-source model/data could be used and verified by external researchers and, hence, be more trustworthy.''} \\[0.5ex]

    \rowcolor{table-row-color}%
    \textbf{Transparency} 
    & Whether the model release is transparent, including code, training dataset, evaluation dataset, etc. 
    & \textit{``Greater transparency on the models and training data would increase my confidence in the models' accuracy.''} \newline \textit{``As researchers, it is not just sufficient to use the service as a black box. I would like to know more about how the models were trained and what data went into it. It builds trust and expands the knowledge.''} \\[0.5ex]

    \textbf{Availability} 
    &	Whether the model is widely available, released to the public, and has no/low barriers for researchers to access 
    & \textit{``World don't have similar access to AI  that might result in systemic discrimination.''} \newline \textit{``...Using for-profit solutions produces costs that require funding, but open-source options are typically less `ready-to-use' and often require setup and potentially also local/available computing resources.''} \\[0.5ex]

    \rowcolor{table-row-color}%
    \textbf{Accountability} 
    & Who should be responsible/accountable for the model and its use
    & \textit{``I don't think it makes a difference whether I’m using os tech or ChatGPT, it's still my responsibility to diligently check the outputs and the responsibility for any risks is on me as the user.''} \newline \textit{``A paid, close-source product could be less reliable, but there is someone to sue if things go awry.''} \\[0.5ex]

    \textbf{Privacy} 
    & Whether data is kept private and stored securely
    & \textit{``Open-source entities are more reliable, and transparent. The risk of using copyrighted data is less. The risk of stealing my personal sensitive data is less.''} \newline \textit{``I trust open-source software more than proprietary/commercial corp if I have to handle personal/sensitive data.''} \\[0.5ex]

    \rowcolor{table-row-color}%
    \textbf{Incentives} 
    & What kinds of incentives drive the creation and release of the model
    & \textit{``LLMs are fundamentally dependent on using people's actual creative and artistic work without their consent.  The motives of the LLM's creator have no effect on that.  Motives similarly have no effect on LLMs' unreliability.  Motives do not change whether LLMs are accountable for their mistakes, because LLMs cannot be accountable.''} \newline \textit{``I can't trust the objectives and implementations of a closed commercial corporation.''} \\[0.5ex]
    
    \textbf{Neutrality} 
    & Whether tools are considered neutral and who created/owns the tool does not matter 
    & \textit{``The technology is the same (whether it's provided by a commercial or an open source entity), so my perceptions are the same.''} \newline \textit{``I don’t know enough about what’s going on (or not) `behind the scenes' in LLMs regardless of where they're from—my perception of them remains that they’re unethical and not useful.''} \\[0.5ex]

    \rowcolor{table-row-color}%
    \textbf{Ethics \& Bias} 
    & Whether the model is biased or uses unethical methods and data
    & \textit{``...a for-profit company might produce models which are favouring a certain kind of opinion, line of thought, or products. These might be equivalently good/bad to a non-profit model, but poses this inherent bias by whomever is currently financially invested into the entity''} \newline \textit{``The fundamental issues of bias, hallucination, ethical issues of invention, the need for review, etc. will not change whether the LLM is open-source or closed.''}  \\[0.5ex]

    \textbf{Quality} 
    & Whether the model has good performance (outputs are useful and of high quality)
    & \textit{``I´d rather use open-source LLMs, but I acknowledge that their performance is still behind commercial models which makes it difficult to use them.''} \newline \textit{``I would be concerned that the commercial aspects would affect the results served. Perhaps we would be directed more to paid sources.''}\\[0.5ex]
    
    \bottomrule
    \end{tabular}
    \caption{Themes found in \textbf{Q64:} \textit{Would your perceptions of the benefits and risks of using LLMs be affected by whether the LLM is part of \textbf{an open-source or non-profit entity versus a commercial corporation}? Why or why not?}} 
    \label{table:qual-themes-64}
\end{table}

%% file: tables/qual_themes_65.tex
\begin{table}[H]
    \centering
    \footnotesize
    \begin{tabular}{@{}p{2cm}|p{4cm}p{8cm}@{}}
    \toprule
    \textbf{Theme} & \textbf{Description} & \textbf{Example} \\
    \midrule
    
    \textbf{Language Equity} 
    & Bridging language gaps to support non-native English speakers
    & \textit{``I am not a native English speaker, so LLMs help me with the language barrier. ''} \newline \textit{``Because I'm not Native American, I've received a number of negative comments from reviewers, and I've always wondered how I can write like a Native American if I'm not one. Today, with the LLM tools, I can understand the terms I use that aren't Native American, and I'm able to improve my writing.''} \\[0.5ex]

    \rowcolor{table-row-color}%
    \textbf{Other Equity} 
    & Removing barriers for researchers (not language), such as supporting neuro-divergent researchers, researchers with limited resources, or researchers without programming experience 
    & \textit{``For researchers without programming skills, LLMs allow data analysis without programming. What they need to learn is how to make good natural language prompts.''} \newline \textit{``Improve understandability for non-specialists''} \newline
    \textit{``For honest researchers in resource-constrained developing countries, with little to no research funding, availability and use of LLMs is a game-changer leveling the playing field with other researchers in more fortunate climes. ''} \\[0.5ex]

    \textbf{Efficiency} 
    & Saving time and resources during the research process
    & \textit{``Enhancing work efficiency and the capability of information gathering and extraction, thereby enabling researchers to focus more energy on creative tasks.''} \newline \textit{``Speeding up the academic writing/reading will speed up the research cycle of the community.''} \\[0.5ex]

    \rowcolor{table-row-color}%
    \textbf{Routine Task \newline Assistance} 
    & Completing routine or repetitive tasks, freeing the researcher to focus on higher level tasks
    & \textit{``I think the main benefit could be saving time that would otherwise have to be spent on relatively 'mechanical' tasks, such as literature search, writing code for data cleaning and analysis, creating nice figures for publication and similar.''} \newline \textit{``Anything repetitive would benefit from LLMs''} \\[0.5ex]

    \textbf{Search} 
    & Search and information retrieval tasks, which might include literature review 
    & \textit{``Information seeking actions (like search) seem to often depend on serial reformulations of concepts, rephrasings, etc. to try to find the right thing. Even if LLMs hallucinate sources, this can be very helpful in those intermediary steps that are part of searching for things. There's just too much out there.''} \newline \textit{``LLMs can give you enough information to continue searching online.''} \\[0.5ex]

    \rowcolor{table-row-color}%
    \textbf{Literature Review} & Helping with literature-related tasks, such as finding related work, writing literature reviews, and summarizing literature & \textit{``Easing the otherwise time-consuming and burdensome processes in research. Tasks like lit searching, reading through content to determine study strengths/weaknesses/biases, extracting data from pubs...''} \newline \textit{``The internet is one big pile of noise. Publications are a somewhat less noisy, but still rather noisy pile of information. LLM's cut through that to some degree.''} \\[0.5ex]

    \textbf{Editing} & Editing tasks, such as rephrasing sentences & \textit{``Mostly rephrasing, rewriting, condensation, bulleting''} \newline \textit{``Editing and language perfection, something like an advanced Grammarly.''} \\[0.5ex]

    \rowcolor{table-row-color}%
    \textbf{Overcoming \newline Writer's Block} & Helping researchers to begin writing (to put something on the blank page) & \textit{``The major benefit of using LLMs comes in action when we are stuck, for example not knowing the exact word or term, or not finding the answer to some of the ideas about how it can be used or how it can be applied.''} \newline \textit{``...loosing the fear of a blank sheet of paper when you don`t know how to start your article...''} \\[0.5ex]

    \textbf{Broadening \newline Perspectives} & Helping researchers discover perspectives and diversify their sources & \textit{``LLM are remarkable important as they reduce generic data and improve novelty of work''} \newline \textit{``Always available ”colleague” to discuss ideas with and get feeback from.''} \\[0.5ex]

    \rowcolor{table-row-color}%
    \textbf{Programming} & Supporting programming tasks, including debugging and writing new code & \textit{``I basically code using CoPilot + GPT now, often for research glue code.''} \newline \textit{``Definitely saves time on coding applications where I'm not aware of certain routines or packages (for example python packages that already exist). Very helpful to get syntax correctly. Get feedback on coding errors...''} \\[0.5ex]

    \textbf{Brainstorming} & Helping brainstorm, organize ideas, and get feedback on ideas & \textit{``LLMs are a great tool to help you create hypotheses, as a way to brainstorm, where there really are no wrong ideas and therefore you cannot suffer with any potential misleading information, as you are expected to have domain expertise anyway.''} \newline \textit{``Having an always available writing companion to help with ideation, divergent thinking, encouragement, and general advice.''} \\[0.5ex]
    
    \bottomrule
    \end{tabular}
    \caption{Themes found in \textbf{Q65:} \textit{Given your ratings above on the benefits/usefulness of these different ways of using LLMs, can you explain what you see to be the main \textbf{benefits/usefulness} for the academic community and for you as a researcher?}} 
    \label{table:qual-themes-65}
\end{table}

%% file: tables/qual_themes_66.tex
\begin{table}[H]
    \centering
    \footnotesize
    \begin{tabular}{@{}p{2cm}|p{3cm}p{9cm}@{}}
    \toprule
    \textbf{Theme} & \textbf{Description} & \textbf{Example} \\
    \midrule

    \textbf{Hallucination \newline \& Misinformation} 
    & Production and spread of incorrect information invented by the model
    & \textit{``Sometimes it creates so complicated hallucinations so that even an expert can think that what it writes it true although it is not.''} \newline \textit{``Putting more falsehoods into [the internet's] shared memory is a crime.''} \\[0.5ex]

    \rowcolor{table-row-color}%
    \textbf{Inaccuracy} 
    & Incorrect conclusions and analyses
    & \textit{``There is a risk of less experienced scientists using these technologies as they are unable to check if the outputs are correct as easily as someone with more experience/intuition.''} \newline \textit{``The risks are proportional to prior knowledge of the subject.''} \\[0.5ex]
    
    \textbf{Biases} 
    & Model's outputs could contain biases and stereotypes
    & \textit{``Promotions of some papers more than others - further marginalizing voices of those who are already less discovered and less cited despite similar quality of papers.''} \newline \textit{``I worry about biased LLMs influencing the research directions we choose and the conclusions we draw.''} \\[0.5ex]

    \rowcolor{table-row-color}%
    \textbf{Lack of Disclosure} 
    & Attribution or disclosure of LLM usage
    & \textit{``Not acknowledging the use of the AI - universities have totally different policies.''} \newline \textit{``The advancement of this technology will make these models a kind of co-author, to the point where we will not know the real contribution of the human component.''} \\[0.5ex]

    \textbf{Plagiarism} 
    & Risks of plagiarism when using models to generate paper text
    & \textit{``Blind trust in a system that is hard to understand which could lead to accidental plagiarism as it's not easy to understand which information LLMs base their outputs on.''} \newline \textit{``plagiarism at scale, the community doesn't have enough time to check all the existing papers''} \\[0.5ex]

    \rowcolor{table-row-color}%
    \textbf{Disrespecting \newline Authorship} 
    & Copyright and other concerns related to ownership of training data and model outputs
    & \textit{``...its use to profit off of text whose authors weren't compensated is pretty fucked up.''} \newline \textit{``There are issues of integrity where the data that is trained on doesn't necessarily belong to the model makers, and the model's output doesn't belong to the user.''} \\[0.5ex] 

    \textbf{Fabrication} 
    & Using LLMs to fabricate data and research results
    & \textit{``The risk of reporting `results' based on synthetic data without actually having conducted any experiment.''} \newline \textit{``LLMs are tools for automated plagiarism and data fabrication that pose an existential threat to the network of trust essential for the integrity of academic work and the proper attribution of credit.''} \\[0.5ex]

    \rowcolor{table-row-color}%
    \textbf{Decreasing \newline Creativity} 
    & Effects of model use on the creativity and originality of research work 
    & \textit{``...llms are going to make people less creative over time. this needs of course more thinking and evidence, but to me, ppl may not start thinking or collaborating human to human to find valuable H2H collaborative ideas, but rather M2H ideas, which might miss the human touch.''} \newline \textit{``The main general risk is to flatten on "average", which is the worst thing that may happen for a researcher (and it is already happening for arts such music, since this would block innovation.''} \\[0.5ex]

    \textbf{Pollution of \newline Research \newline Ecosystem} 
    & Decreasing quality of research that leads to overall pollution of information and community
    & \textit{``The huge number of poor quality papers out there are already making science more difficult, and I can see LLMs making this much worse.''} \newline \textit{``We need better judgment, slower science, and more thoughtful and ambitious work right now, not the opposite. Otherwise, we risk ridding science of its most special attributes just to crank out more papers.''} \\[0.5ex]

    \rowcolor{table-row-color}%
    \textbf{Decreasing \newline  Diligence} 
    & Over-reliance on and trust in models leading to decreasing diligence of researchers
    & \textit{``Speed, copy/paste attitude, less research propositions, less reading of the full article, not enough training...''} \newline \textit{``The main risks are related to human tendency to be lazy and appreciate convenience too much. It would be easy to overtrust LLM output the inherent opacity of the technology invites people to do so. It is hard to check why particular output is produced and people have a natural talent to explain and justify things, even completely wrong things when it is more convenient.''} \\[0.5ex]

    \textbf{Deskilling} 
    & Loss of research skills due to reliance on models
    & \textit{``...sort of like with self-driving cars which, if we ever get ones that work a bit, people will stop learning how to drive, which will be bad when the things actually get confused and hand you the wheel!''} \newline \textit{``Learning how to relay information is an important skill that must be cultivated throughout ones career.  The process of writing the information gives you the way to check whether what you have done actually makes sense and provides value.''} \\[0.5ex]
    
    \bottomrule
    \end{tabular}
    \caption{Themes found in \textbf{Q66:} \textit{Given your ratings above on the risks/ethical considerations of these different ways of using LLMs, can you explain what you see to be the main \textbf{risks/ethical considerations} for the academic community and for you as a researcher?}} 
    \label{table:qual-themes-66}
\end{table}